\documentclass[runningheads]{llncs}
\usepackage[T1]{fontenc}
\usepackage{array}
\usepackage{booktabs}
\usepackage{multirow}
\usepackage[table,xcdraw]{xcolor}
\usepackage{graphicx}
\usepackage{marvosym}
\usepackage{amsmath, amssymb}
\newcommand{\textred}{\textcolor{red}}
\newcommand{\textblue}{\textcolor{blue}}

\newcommand{\ck}{\checkmark}

\begin{document}
\title{YONA: You Only Need One Adjacent Reference-frame for Accurate and Fast Video Polyp Detection}
\titlerunning{YONA for Accurate and Fast Video Polyp Detection}

\author{
    Yuncheng Jiang\inst{1,2,4,\star}  \and                     
    Zixun Zhang\inst{1,2,4} \thanks{Equal contribution} \and   
    Ruimao Zhang\inst{3} \and                                  
    Guanbin Li\inst{5} \and                                  
    Shuguang Cui\inst{1,2} \and                                
    Zhen Li\inst{1,2,4}                                        
\textsuperscript{\Letter}}

\authorrunning{Y. Jiang et al.}

\institute{
SSE, The Chinese University of Hong Kong, Shenzhen
\and
FNii, The Chinese University of Hong Kong, Shenzhen
\and
SDS, The Chinese University of Hong Kong, Shenzhen
\and
Shenzhen Research Insititute of Big Data
\and
School of Computer Science and Engineering, Sun Yat-sen University\\
\email{yunchengjiang@link.cuhk.edu.cn, lizhen@cuhk.edu.cn}}

\maketitle

\begin{abstract}
Accurate polyp detection is essential for assisting clinical rectal cancer diagnoses. Colonoscopy videos contain richer information than still images, making them a valuable resource for deep learning methods. However, unlike common fixed-camera video, the camera-moving scene in colonoscopy videos can cause rapid video jitters, leading to unstable training for existing video detection models. In this paper, we propose the \textbf{YONA} (\textbf{Y}ou \textbf{O}nly \textbf{N}eed one \textbf{A}djacent Reference-frame) method, an efficient end-to-end training framework for video polyp detection. YONA fully exploits the information of one previous adjacent frame and conducts polyp detection on the current frame without multi-frame collaborations. Specifically, for the foreground, YONA adaptively aligns the current frame's channel activation patterns with its adjacent reference frames according to their foreground similarity. For the background, YONA conducts background dynamic alignment guided by inter-frame difference to eliminate the invalid features produced by drastic spatial jitters. Moreover, YONA applies cross-frame contrastive learning during training, leveraging the ground truth bounding box to improve the model's perception of polyp and background. Quantitative and qualitative experiments on three public challenging benchmarks demonstrate that our proposed YONA outperforms previous state-of-the-art competitors by a large margin in both accuracy and speed.


\end{abstract}

\section{Introduction}
\begin{figure}[t]
    \centering
    \includegraphics[width=0.9\textwidth]{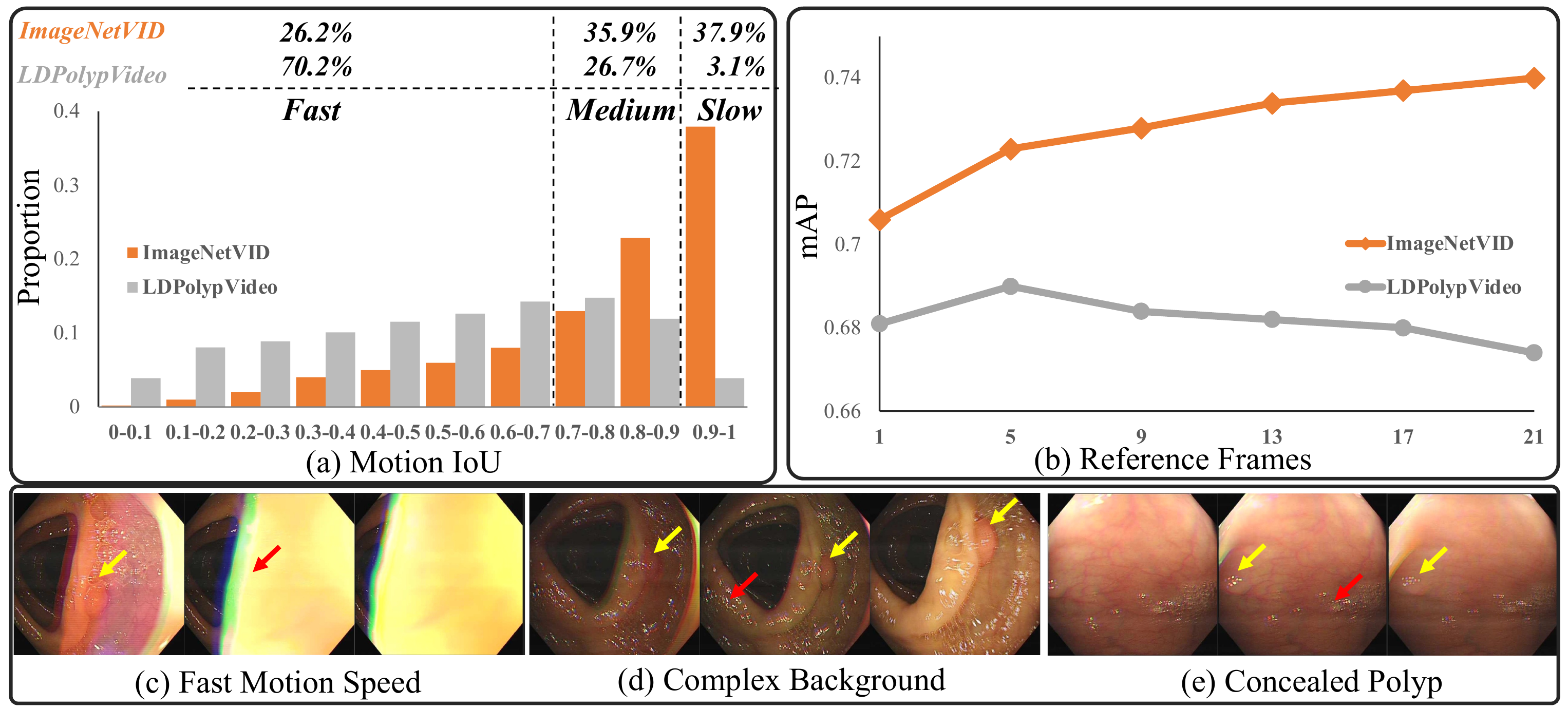}
    \caption{
    (a) The histogram of the motion IoUs distribution on two datasets. Lower motion IoU denotes a faster target moving speed. The proportion of slow, medium and fast-moving targets is listed at the top of the figure.
    (b) The performance of FGFA~\cite{fgfa} using multiple reference frames increases on ImageNetVID while decreasing on LDPolypVideo.
    (c) The typical challenges in colonoscopy videos. Yellow arrows point to the polyp, and red arrows point to distraction that causes false detection.}
    \label{fig:teaser}
\end{figure}

Colonoscopy plays a crucial role in identifying and removing early polyps and reducing mortality rates associated with rectal cancer. Over the past few years, the research community has devoted great effort to understanding colonoscopy videos using either optical flow~\cite{OptCNN,AIPDT} or temporal information aggregation~\cite{qadir2019improving,tajbakhsh2015automatic,stft,2d3d} between multiple frames.

However, those works are mainly designed based on the experience of previous natural video object detection studies, ignoring the inherent uniqueness of the colonoscopy motion patterns. Thus, we rethink the video polyp detection task and conclude three core challenges in colonoscopy videos. \textbf{1) Fast motion speed}. In Fig.~\ref{fig:teaser}(a), we show the target motion speed~\cite{fgfa} $\footnote{averaged intersection-over-union scores of target in the nearby frames ($\pm 10$ frames)}$ on ImageNetVID~\cite{ILSVRC15} (natural) and LDPolypVideo~\cite{ldpolypvideo} (colonoscopy) dataset. The motion speed in ImageNetVID evenly distributes in three intervals. In contrast, most targets in LDPolypVideo fall in the fast speed zone, leading to a large variance in the adjacent foreground features, like motion blur or occlusion, as shown in Fig.~\ref{fig:teaser}(c). Thus we conjecture that collaborating too many frames for polyp video detection will increase the misalignment between adjacent frames and leads to poor detection performance. Fig.~\ref{fig:teaser}(b) shows the performance of FGFA~\cite{fgfa} on two datasets with increasing reference frames. The different trends of the two lines confirm our hypothesis. \textbf{2) Complex background} Different from the common camera-fixed videos, the camera-moving of colonoscopy video will introduce large disturbances between adjacent frames (e.g., specular reflection, bubbles, water, etc.), as shown in Fig.~\ref{fig:teaser}(d). Those abnormalities disrupt the integrity of background structures and thus affect the effect of multi-frame fusion. \textbf{3) Concealed polyps} As shown in Fig.~\ref{fig:teaser}(e), we noticed that some polyps could be seen as concealed objects in the colonoscopy video since such polyps have a very similar appearance to the intestine wall. The model will be confused by such frames in inference and result in high false-positive or false-negative predictions. 

To address the above issues, we propose the \textbf{YONA} framework, which fully exploits the reference frame information and only needs one adjacent reference frame for accurate video polyp detection. Specifically, we propose the Foreground Temporal Alignment (FTA) module to explicitly align the foreground channel activation patterns between adjacent features according to their foreground similarity. In addition, we design the Background Dynamic Alignment (BDA) module after FTA that further learns the inter-frame background spatial dynamics to better eliminate the influence of motion speed and increase the training robustness. Finally, parallel to FTA and BDA, we introduce the Cross-frame Box-assisted Contrastive Learning (CBCL) that fully utilizes the box annotations to enlarge polyp and background discrimination in embedding space.

In summary, our contributions are in three-folds: (1) To the best of our knowledge, we are the first to investigate the obstacles to the development of existing video polyp detectors and conclude that two-frame collaboration is enough for video polyp detection. (2) We propose the YONA, a novel framework for video polyp detection. It composes the foreground and background alignment modules to align the features under the fast-moving condition. It further introduces the cross-frame contrastive learning module to enhance the model's discrimination ability of polyps and intestine walls. (3) Extensive experiments demonstrate that our YONA achieves new state-of-the-art performance on three large-scale public video polyp detection datasets.

\section{Method}
\label{sec:method}

\begin{figure}[t]
    \centering
    \includegraphics[width=0.9\textwidth]{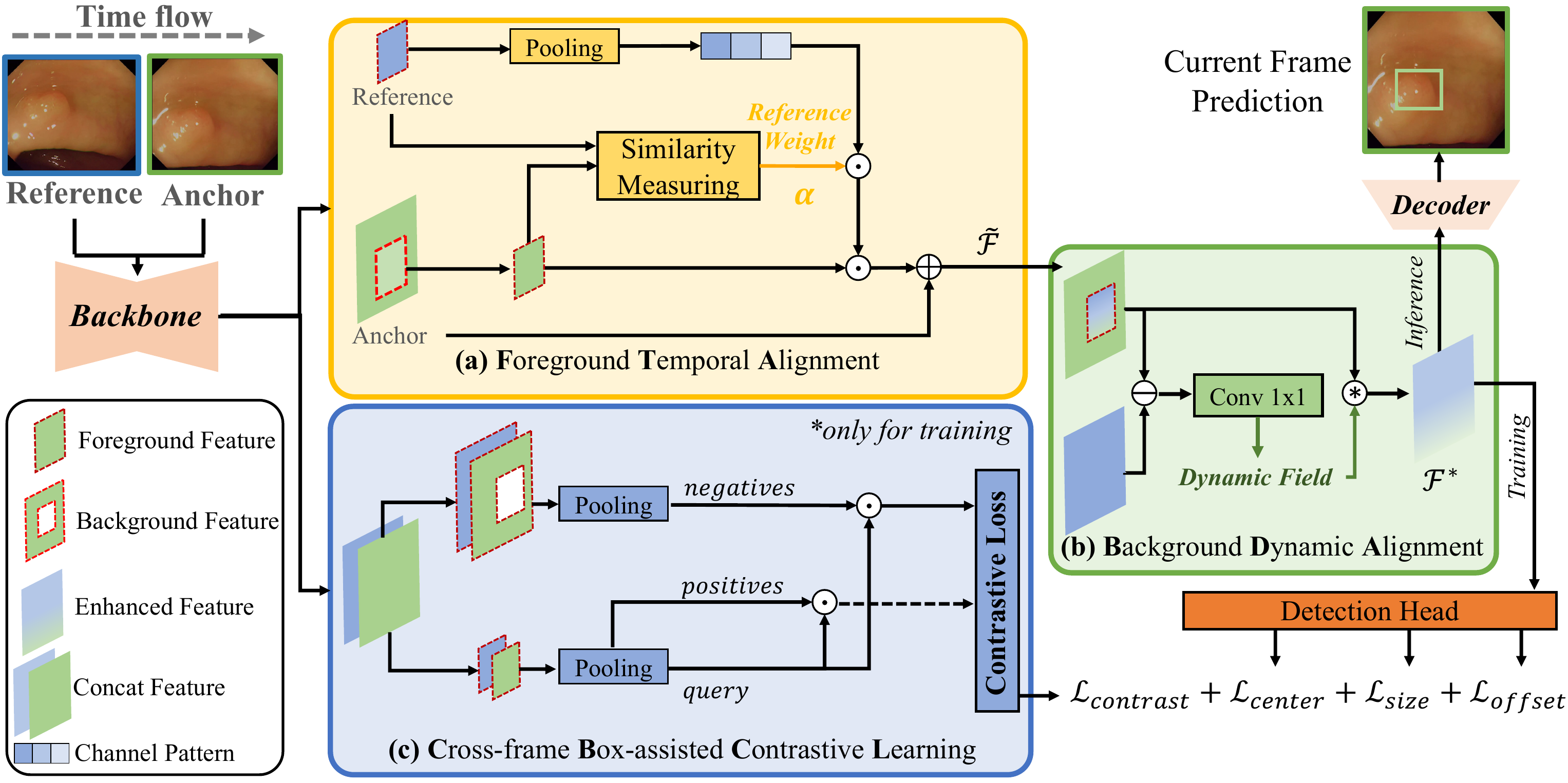}
    \caption{Illustration of our proposed video polyp detection framework, YONA. 
    It first aligns the foreground channel patterns between the anchor and reference frame in (a). 
    Then it extracts polyp context guided by dynamic field in (b). 
    Meanwhile, YONA enhances the discrimination ability via contrastive learning in (c) during training. 
    The final output of (b) is used to predict the bounding box of the current frame.
    }
    \label{fig:framework}
\end{figure}

The whole pipeline is shown in Fig.~\ref{fig:framework}. 
We leverage the CenterNet~\cite{centernet} as the base detector. Given a clip of a colonoscopy video, we take the current frame as anchor $\mathit{I^a}$ and its adjacent previous frame as reference $\mathit{I^r}$. The binary maps $M^a, M^r$ are generated using the bounding box of anchor and reference, where the foreground pixels are assigned with $1$ while the background with $0$. At each step, YONA first extracts multi-scale features from $I^a, I^r$ using the backbone. Then, multi-scale features are fused and up-sampled to the resolution of the first stage as the intermediate features $F^a, F^r$. Then, we conduct foreground temporal alignment (Fig.~\ref{fig:framework}(a)) on intermediate features to align their channel activation pattern. Next, the enhanced anchor feature $\Tilde{F}$ is further refined by the background dynamic alignment module (Fig.~\ref{fig:framework}(b)) to mitigate the rapid dynamic changes in the spatial field. The BDA's output $F^*$ is used to compute the detection loss. Meanwhile, the intermediate features and binary maps are used to calculate the contrastive loss during training to improve the model's perception of polyp and background (Fig.~\ref{fig:framework}(c)).

Overall, the whole network is optimized with the combination loss function in an end-to-end manner. The final loss is composed of the same detection loss with CenterNet and our proposed contrastive loss, formulated as $\mathcal{L} =\mathcal{L}_{\text {detection}}+\lambda_{contrast}\mathcal{L}_{\text {contrast}}$.

\subsection{Foreground Temporal Alignment}
Since the camera moves at a high speed, the changes in the frame are very drastic for both foreground and background targets. As a result, multi-frame (reference>3) fusion may easily incorporate more noise features into the aggregation features. On the other hand, the occluded or distorted foreground context may also influence the quality of aggregation. Thus we propose to conduct temporal alignment between adjacent features by leveraging the foreground context of only \textbf{one} adjacent reference frame. It is designed to align the certain channel's activation pattern of anchor feature to its preceding reference feature. Specifically, given the intermediate features $F^a, F^r$ and reference binary map $M^r$, we first pooling $F^r$ to 1D channel pattern $f^r$ by the binary map on the spatial dimension ($\mathbb{R}^{N \times C \times H \times W} \rightarrow \mathbb{R}^{N \times C \times 1}$) and normalize it to $[0,1]$:
\begin{equation}
\label{eq:1}
\begin{aligned}
    f^r &= \text{norm}\left[\text{Pooling}(F^r)\right] \\
    \text{Pooling}(F^r) &= \text{sum}_{HW}\left[F^r(x,y)]/\text{sum}[M^r(x,y)\right] \quad \text{if}~M^r(x,y)=1
\end{aligned}
\end{equation}
Then, the foreground temporal alignment is implemented by channel attention mechanism, where the attention maps are computed by weighted dot-product. We obtain the enhanced anchor feature by adding the attention maps with the original anchor feature through skip connection to keep the gradient flow.
\begin{equation}
    \Tilde{\mathcal{F}} =  \left[\alpha f^r \odot F^a(x,y)\right] \oplus F^a  \quad \text{if}~M^r(x,y)=1
\end{equation}
where $\alpha$ is the adaptive weight by similarity measuring.

At the training stage, the ground truth boxes of the reference frame are used to generate the binary map $M^r$. 
During the inference stage, we conduct FTA only if the validated bounding box of the reference frame exists, where "validated" denotes the confidence scores of detected boxes are greater than $0.6$.
Otherwise, we will skip this process and feed the original inputs to the next module.
\\

\noindent
\textbf{Adaptive Re-weighting by Similarity Measuring}
As discussed above, due to video jitters, adjacent frames may change rapidly at the temporal level, and directly fusing the reference feature will introduce noisy information and misguide the training. Thus we designed an adaptive re-weighting method by measuring the feature similarity, where the weight indicates the importance of the reference feature to the anchor feature. Specifically, if the foreground feature of the reference is close to the anchor, it is assigned a larger weight at all channels. Otherwise, a smaller weight is assigned. For efficiency, we use the cosine similarity metric~\cite{CosineNU} to measure the similarity, where $f^a$ is the 1D channel pattern of $F^a$ computed with Eq.~\ref{eq:1}:
\begin{equation}
    \alpha =  \mathit{exp} \left(\frac{f^r\cdot f^a}{|f^r||f^a|}\right)
\end{equation}

\subsection{Background Dynamic Alignment}

The traditional convolutional-based object detector can detect objects well when the background is stable. 
However, once it receives obvious interference, such as light or shadow, the background changes may cause the degradation of spatial correlation and lead to many false-positive predictions. Motivated by the inter-frame difference method~\cite{FrameDIff}, we first mine the dynamic field of adjacent background contents, then consult to deformable convolution~\cite{dai2017deformableconv} to learn the inherent geometric transformations according to the intensity of the dynamic field. In practice, given the enhanced anchor feature $\Tilde{\mathcal{F}}$ from FTA and reference feature $F^r$, the inter-frame difference is defined as the element-wise subtraction of enhanced anchor and reference feature. Then a $1 \times 1$ convolution is applied on the difference to generate dynamic field $\mathcal{D}$, which encodes all spatial dynamic changes between adjacent frames.
\begin{equation}
\begin{aligned}
    \mathcal{D}  = \text{Conv}_{1\times1}(\Tilde{\mathcal{F}} - F^r)
\end{aligned}
\end{equation}
Finally, a $3 \times 3$ deformable convolution embeds the spatial dynamic changes of $\mathcal{D}$ on the enhanced anchor feature $\Tilde{F}$.
\begin{equation}
\begin{aligned}
    \mathcal{F}^* &= \text{DeConv}_{3\times3}(\mathcal{\Tilde{F},D})    
\end{aligned}
\end{equation}
where $D$ works as the deformable offset and $F^*$ is the final aligned anchor feature. Then the enhanced anchor feature is fed into three detection heads composed of a $3\times 3$ Conv and a $1\times 1$ Conv to produce center, size, and offset features for detection loss:
\begin{equation}
    \mathcal{L}_{\text{detection}} =\mathcal{L}^{\text {center}}_{\text {focal}}+\lambda_{size}\mathcal{L}^{\text {size}}_{\text {L1}}+\lambda_{off}\mathcal{L}^{\text {offset}}_{\text {L1}}
\end{equation}
where $\mathcal{L}_{\text{focal}}$ is focal loss and $\mathcal{L}_{\text{L1}}$ is L1 loss.

\subsection{Cross-frame Box-assisted Contrastive Learning}
Typically, in colonoscopy videos, some concealed polyps appear very similar to the intestine wall in color and texture. Thus, an advanced training strategy is required to distinguish such homogeneity. Inspired by recent studies on supervised contrastive learning~\cite{wang2021exploring}, we select the foreground and background region on both two frames guided by ground truth boxes to conduct contrastive learning. In practice, Given a batch of intermediate feature maps $F^a, F^r\in\mathbb{R}^{N\times T\times C\times H\times W}$ and corresponding binary maps $M^a, M^r\in\mathbb{R}^{N\times T\times H \times W}$, we first concatenate the anchor and reference at the batch-wise level as $\hat{F}\in\mathbb{R}^{NT\times C\times H\times W}$ and $\hat{M} \in \mathbb{R}^{NT\times H\times W}$ to exploit the cross-frame information. Then we extract the foreground and background channel patterns of cross-frame feature $\hat{F}$ using the Eq.~\ref{eq:1} base on $\hat{M}(x,y)=1$ and $\hat{M}(x,y)=0$, respectively. After that, for each foreground channel pattern, which is the "query", we randomly select another different foreground feature as the "positive", while all the background features in the same batch are taken as the "negatives". Finally, we calculate the one-step contrastive loss by InfoNCE~\cite{wang2021exploring}:
\begin{equation}
    \mathcal{L}_{j}^{\rm NCE} = {-{\rm log}\frac{{\rm exp}(q_j \!\cdot\! i^+ / \tau)}{{\rm exp}(q_j \!\cdot\! i^+/\tau) \!+\! \sum_{i^- \!\in\! \mathcal{N}_j}{{\rm exp}(q_j \!\cdot\! i^- / \tau)}}} \\
\end{equation}
\noindent where $q_j\in\mathbb{R}^{C}, j=0,...,NT$ is the query feature, $i^+\in \mathbb{R}^{C}$ and $i^-\in \mathbb{R}^{NT\times C}$ are positives and negatives.
$ \mathcal{N}_j$ denote embedding collections of the negatives. We repeat this process until every foreground channel pattern is selected and sum all steps as the final contrastive loss:
\begin{equation}
    \label{eq:ct_loss}
    \mathcal{L}_{\text{contrast}} = \frac{1}{NT} \sum_{j=1}^{NT} \mathcal{L}_{j}^{NCE}
\end{equation}

\section{Experiments}
We evaluate the proposed method on three public video polyp detection benchmarks: SUN Colonoscopy Video Database~\cite{suncolon,sunwebsite} (train set: 19,544 frames, test set: 12,522 frames), 
LDPolypVideo~\cite{ldpolypvideo} (train set: 20,942 frames, test set: 12,933 frames), and CVC-VideoClinicDB~\cite{bernal2018polyp} (train set: 7995 frames, test set: 2030 frames). For the fairness of the experiments, we keep the same dataset settings for YONA and all other methods.

We use ResNet-50~\cite{he2016deep} as our backbone and CenterNet~\cite{centernet} as our base detector. Following the same setting in CenterNet, we set $\lambda_{size}=0.1$ and $\lambda_{off}=1$. We set $\lambda_{contrast}=0.3$ by ablation study. Detailed results are listed in the supplement. We randomly crop and resize the images to $ 512 \times 512 $ and normalize them using ImageNet settings. Random rotation and flip with probability $p=0.5$ are used for data augmentation. We set the batch size $N=32$. Our model is trained using the Adam optimizer with a weight decay of $5\times10^{-4}$ for $64$ epochs. The initial learning rate is set to $10^{-4}$ and gradually decays to $10^{-5}$ with cosine annealing. All models are trained with PyTorch~\cite{AdamPaszke2019PyTorchAI} framework. The training setting of other competitors follows the best settings given in their paper.

\begin{table}[t]
\centering
\caption{Performance comparison with other image/video-based detection models. 
P, R, and F1 denote the precision, recall, and F1-score. $\dagger$: results from the original paper with the same data division. The best score is marked as \textred{red}, while the second best score is marked as \textblue{blue}.}
\label{tab:performace}
\renewcommand\arraystretch{1}
\renewcommand\tabcolsep{4pt}
\begin{tabular}{r|ccc|ccc|ccc|c}
    \toprule[1.5pt]
      \multirow{2}{*}{Methods}  & \multicolumn{3}{c|}{SUN Database}  & \multicolumn{3}{c|}{LDPolypVideo} & \multicolumn{3}{c|}{CVC-VideoClinic}  & \multirow{2}{*}{FPS} \\
       & P & R & F1 & P & R & F1 & P & R & F1 & \\ 
    \hline
    Faster-RCNN~\cite{ren2015faster}   & 77.2 & 69.6 & 73.2 & 68.8 & 46.7 & 55.6 & 84.6 & \textred{98.2} & 90.9 & 44.7\\
    FCOS~\cite{tian2019fcos}        & 75.7 & 64.1 & 69.4 & 65.1 & 46.0 & 53.9 & 92.1 & 74.1 & 82.1 & 42.0\\
    CenterNet~\cite{centernet}  & 74.6 & 65.4 & 69.7 & 70.6 & 43.8 & 54.0 & 92.0 & 80.5 & 85.9 & \textred{51.5}\\
    Sparse-RCNN~\cite{sun2021sparse}   & 75.5 & \textblue{73.7} & 74.6 & 71.6 & 47.9 & 57.4 & 85.1 & 96.4 & 90.4 & 40.0\\
    DINO~\cite{zhang2022dino}       & \textblue{81.5} & 72.3 & 76.6 & 68.3 & \textblue{51.1} & 58.4 & \textblue{93.1} & 89.3& 91.2 & 23.0\\
     \hline
    FGFA~\cite{fgfa}          & 78.9 & 70.4 & 74.4 & 68.8 & 48.9 & 57.2 & \textred{94.5} & 89.2 & 91.7 & 1.8\\  
    OptCNN$\dagger$~\cite{OptCNN}   & - & - & - & - & - & - & 84.6 & \textblue{97.3} & 90.5 & -\\  
    AIDPT$\dagger$~\cite{AIPDT}     & - & - & - & - & - & - & 90.6 & 84.5 & 87.5 & -\\  
    MEGA~\cite{chen2020memory}      & 80.4 & 71.6 & 75.7 & 69.2 & 50.1 & 58.1 & 91.6 & 87.7 & 89.6 & 8.1\\
    TransVOD~\cite{zhou2022transvod}    & 79.3 & 69.6 & 74.1 & 69.2 & 49.2 & 57.5 & 92.1 & 91.4 & 91.7 & 8.4\\   
    STFT~\cite{stft}        & \textblue{81.5} & 72.4 & \textblue{76.7} & \textblue{72.1} & 50.4 & \textblue{59.3} & 91.9 & 92.0 & \textblue{92.0} & 12.5\\  
    \textbf{Ours-YONA}             & \textred{83.3} & \textred{74.9} & \textred{78.9} & \textred{75.4} & \textred{53.1} & \textred{62.3} & 92.8 & 93.8 & \textred{93.3} & \textblue{46.3}\\
    \bottomrule[1.5pt]
\end{tabular}
\end{table}

\subsection{Quantitative and Qualitative Comparison}
\noindent
\textbf{Quantitative Comparison}
The comparison results are shown in Tab.~\ref{tab:performace}. Following the standard of \cite{bernal2018polyp}, the Precision, Recall, and F1-scores are used for evaluation. Firstly, compared with the CenterNet baseline, our YONA with three novel designs significantly improved the F1 score by $9.2\%$, $8.3\%$, and $7.4\%$ on three benchmarks, demonstrating the effectiveness of the model design. Besides, YONA achieves the best trade-off between accuracy and speed compared with all other image-based SOTAs across all datasets. Second, for video-based competitors, previous video object detectors with multiple frame collaborations lack the ability for accurate detection on challenging datasets. Specifically, YONA surpasses the second-best STFT~\cite{stft} by $2.2\%$, $3.0\%$, and $1.3\%$ on F1 score on three datasets and $33.8$ on FPS. All the results confirm the superiority of our proposed framework for accurate and fast video polyp detection.

\begin{figure}[t]
    \centering
    \includegraphics[width=0.95\textwidth]{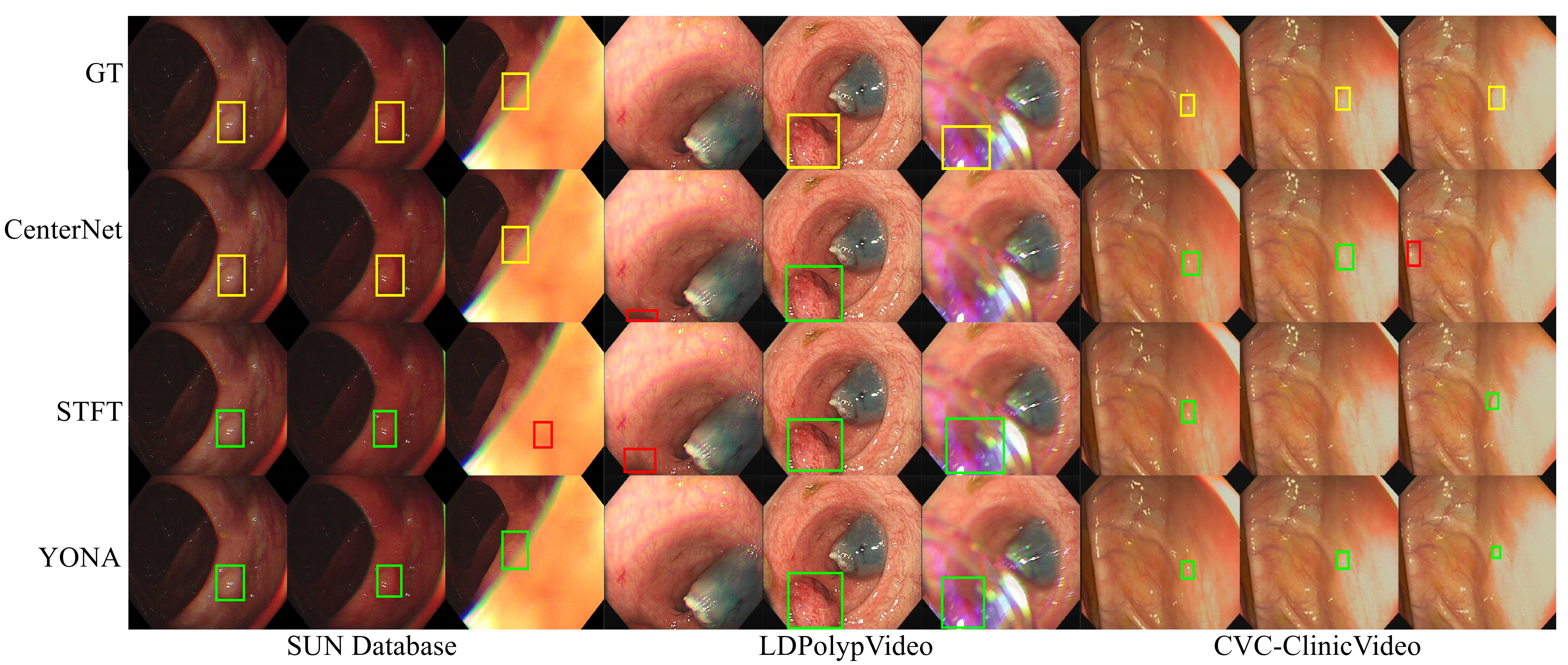}
    \caption{Qualitative results of polyp detection on some video clips. The yellow, green, and red denote the ground truth, true positive, and false positive, respectively} \label{fig:vis}
\end{figure}

\noindent
\textbf{Qualitative Comparison}
Fig.~\ref{fig:vis} visualizes the qualitative results of YONA with other competitors~\cite{centernet,stft}. Thanks to this one-adjacent-frame framework, our YONA can not only prevent the false positive caused by part occlusion (1st and 2nd clips) but also capture useful information under severe image quality (2nd clip). Moreover, our YONA shows robust performance even for challenging scenarios like concealed polyps (3rd clip).

\begin{table}[t]
\centering
\caption{Ablation studies of YONA under different settings. Ada means the adaptive re-weighting by similarity measuring; CW denotes the channel-wise attention~\cite{fu2019dual}; CA denotes the channel-aware attention~\cite{stft}.}
    \label{tab:ablation-modules}
    \renewcommand\arraystretch{1}
    \renewcommand\tabcolsep{4.5pt}
    \begin{tabular}{cccccc|cccc}
    \toprule[1.5pt]
    FTA & CW~\cite{fu2019dual} & CA~\cite{stft} &Ada & BDA  & CBCL & Precision & Recall & F1 & FPS   \\ \hline
        &   &   &     &   &       & 74.6 & 65.4 & 69.7 & \textbf{51.5} \\ \hline
    \ck &   &   &     &   &       & 74.0 & 63.9 & $68.6_{\downarrow1.1}$ & 49.7 \\
    \ck &   &   & \ck &   &       & 80.9 & 70.1 & $75.1_{\uparrow5.4}$ & 48.5 \\
        &\ck&   & \ck &   &       & 78.0 & 65.2 & $71.1_{\uparrow1.4}$ & 48.3 \\
        &   &\ck& \ck &   &       & 80.4 & 68.4 & $73.9_{\uparrow4.2}$ & 45.2 \\  
    \ck &   &   & \ck & \ck &     & 82.0 & 72.2 & $76.8_{\uparrow7.1}$ &  46.3 \\  
    \ck &   &   & \ck & \ck & \ck & \textbf{83.3} & \textbf{74.9} & $\mathbf{78.9}_{\uparrow9.2}$ & 46.3 \\ \bottomrule[1.5pt]
    \end{tabular}
\end{table}

\subsection{Ablation Study} 

We investigated the effectiveness of each component in YONA on the SUN database, as shown in Tab.~\ref{tab:ablation-modules}. It can be observed that all the modules are necessary for precise detection compared with the baseline results. Due to the large variance of colonoscopy image content, the F1 score slightly decreases if directly adding FTA without the adaptive re-weighting strategy. Adding the adaptive weight greatly improves the F1 score by $5.4$. Moreover, we use other two mainstream channel attention mechanisms to replace our proposed FTA for comparison. Compared with them, our FTA with adaptive weighting achieves the largest gain over the baseline and higher FPS. Overall, by combining all the proposed methods, our model can achieve new state-of-the-art performance.

\section{Conclusion}

Video polyp detection is a currently challenging task due to the fast-moving property of colonoscopy video. In this paper, We proposed the YONA framework that requires only one adjacent reference frame for accurate and fast video polyp detection. To address the problem of fast-moving polyps, we introduced the foreground temporal alignment module, which explicitly aligns the channel patterns of two frames according to their foreground similarity. For the complex background content, we designed the background dynamic alignment module to mitigate the large variances by exploiting the inter-frame difference. Meanwhile, we employed a cross-frame box-assisted contrastive learning module to enhance the polyp and background discrimination based on box annotations. Extensive experiment results confirmed the effectiveness of our method, demonstrating the potential for practical use in real clinical applications.

\section*{Acknowledgements} 
This work was supported in part by Shenzhen General Program No. JCYJ202205\\30143600001, by the Basic Research Project No. HZQB-KCZYZ-2021067 of Hetao Shenzhen HK S$\&$T Cooperation Zone, by Shenzhen-Hong Kong Joint Funding No. SGDX20211123112401002, NSFC with Grant No. 62293482, by Shenzhen Outstanding Talents Training Fund, by Guangdong Research Project No. 2017ZT07X152 and No. 2019CX01X104, by the Guangdong Provincial Key Laboratory of Future Networks of Intelligence (Grant No. 2022B1212010001), by the Guangdong Provincial Key Laboratory of Big Data Computing, The Chinese University of Hong Kong, Shenzhen, by the NSFC 61931024$\&$81922046, by  the Shenzhen Key Laboratory of Big Data and Artificial Intelligence (Grant No. ZDSYS201707251409055), and the Key Area R$\&$D Program of Guangdong Province with grant No. 2018B030338001, by zelixir biotechnology company Fund, by Tencent Open Fund.

\bibliographystyle{splncs04}
\bibliography{paper235}

\end{document}


\title{YONA: You Only Need One Adjacent Reference-frame for Accurate and Fast Video Polyp Detection\\
Supplementary Material
}

\author{Yuncheng Jiang\inst{1,2,4,\star}
\and
Zixun Zhang\inst{1,2,4}\thanks{Equal contribution}
\and
Ruimao Zhang\inst{3}
\and 
Guanbin Li\inst{5}
\and
Shuguang Cui\inst{1,2}
\and Zhen Li\inst{1,2,4}\textsuperscript{\Letter}}
%
\authorrunning{A. Author et al.}
%
\institute{
SSE, The Chinese University of Hong Kong, Shenzhen
\and
FNii, The Chinese University of Hong Kong, Shenzhen
\and
SDS, The Chinese University of Hong Kong, Shenzhen
\and
Shenzhen Research Insititute of Big Data
\and
School of Computer Science and Engineering, Sun Yat-sen University\\
\email{lizhen@cuhk.edu.cn}}
%
%

\maketitle
%

\begin{table}[ht]
    \centering
    \caption{Impact of $\lambda_{\text{contrast}}$ on F1-score.}
    \renewcommand\arraystretch{1}
    \renewcommand\tabcolsep{5pt}
    \begin{tabular}{c|c|c|c}
    \toprule[1.5pt]
    $\lambda_{\text{contrast}}$ & SUN Database   & LDPolypVideo & CVC-VideoClinic   \\ \cmidrule(l){1-4} 
    0.1  & 78   & 61.7 & 92.6         \\
    0.2  & 78.5 & \textbf{62.3} & 93.4         \\  
    \textbf{0.3}  & \textbf{78.9} & \textbf{62.3} & \textbf{93.4}         \\  
    0.4  & 78.6 & 62.1   & 93.3         \\ 
    0.5  & 78.2 & 62   & 93.1         \\ 
    0.6  & 78   & 61.8 & 93           \\ 
    0.7  & 78.1 & 61.5 & 92.6         \\ 
    0.8  & 78.3 & 61.2 & 92.4         \\ 
    0.9  & 77.9 & 61.6 & 92.6         \\ 
    1    & 77.7 & 61.7 & 92.8         \\ 
    \bottomrule[1.5pt]
    \end{tabular}
\end{table}

\begin{table}[ht]
    \centering
    \caption{Impact of the reference frame number on accuracy (F1-score) and inference speed (FPS).}
    \renewcommand\arraystretch{1}
    \renewcommand\tabcolsep{5pt}
    \begin{tabular}{c|ccccccc}
    \toprule[1.5pt]
    \multirow{2}{*}{Dataset} & \multicolumn{7}{c}{\# Reference Frames}\\ 
                    & 0    & 1*   & 2    & 4    &  6   &  8   & 10   \\\hline
    SUN Database    & 69.7 & 78.9 & 78.1 & 76.4 & 75.6 & 73.7 & 73.0 \\
    LDPolypVideo    & 54.0 & 62.3 & 61.1 & 59.7 & 56.6 & 55.1 & 54.5 \\
    CVC-VideoClinic & 85.9 & 93.3 & 92.6 & 91.2 & 89.8 & 87.5 & 86.4 \\\hline
    FPS             & 51.5 & 46.3 & 33.6 & 26.5 & 18.9 & 13.4 & 10.8 \\
    \bottomrule[1.5pt]
    \end{tabular}
\end{table}

\begin{figure}[t]
    \centering
    \includegraphics[width=0.9\textwidth]{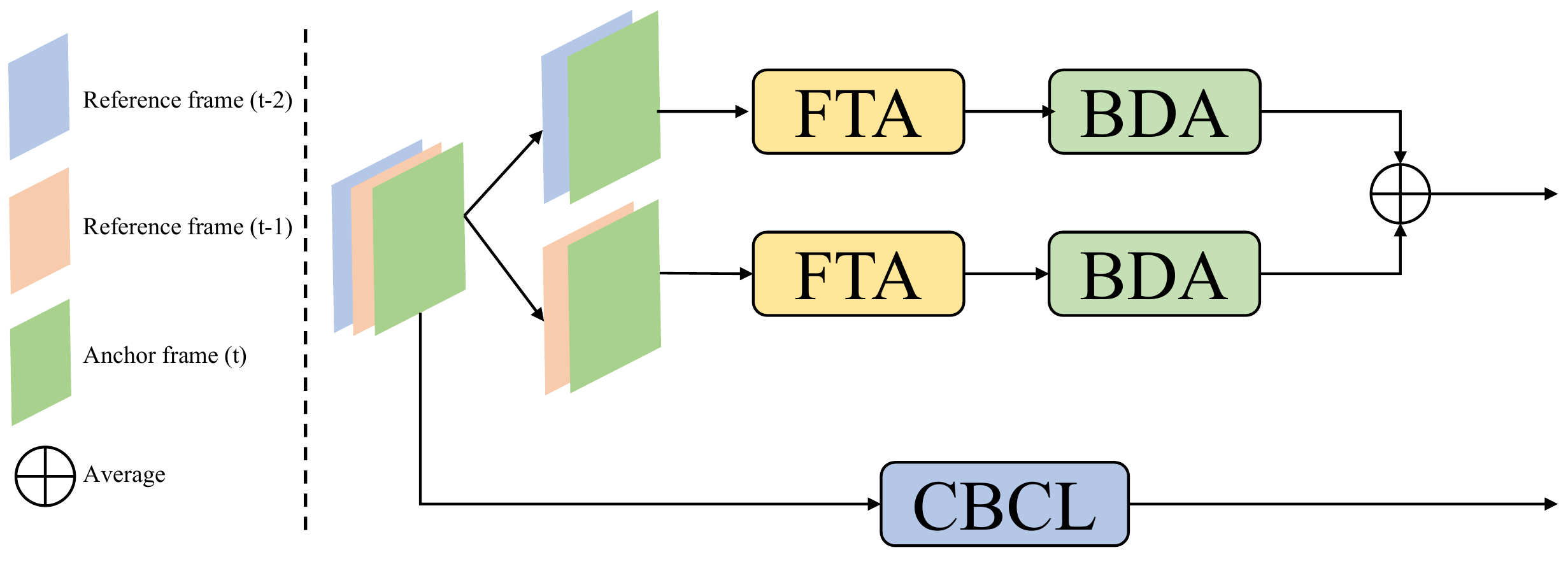}
    \caption{Illustration of YONA framework with multiple reference frames. For simplicity, here we take two reference frames as an example. It's noted that the design of multi-frame architecture is not the main concern and contribution of our work. Thus, we just adopt the naive average fusion strategy.}
    \label{fig:supplement}
\end{figure}

\begin{figure}[t]
    \centering
    \includegraphics[width=\textwidth]{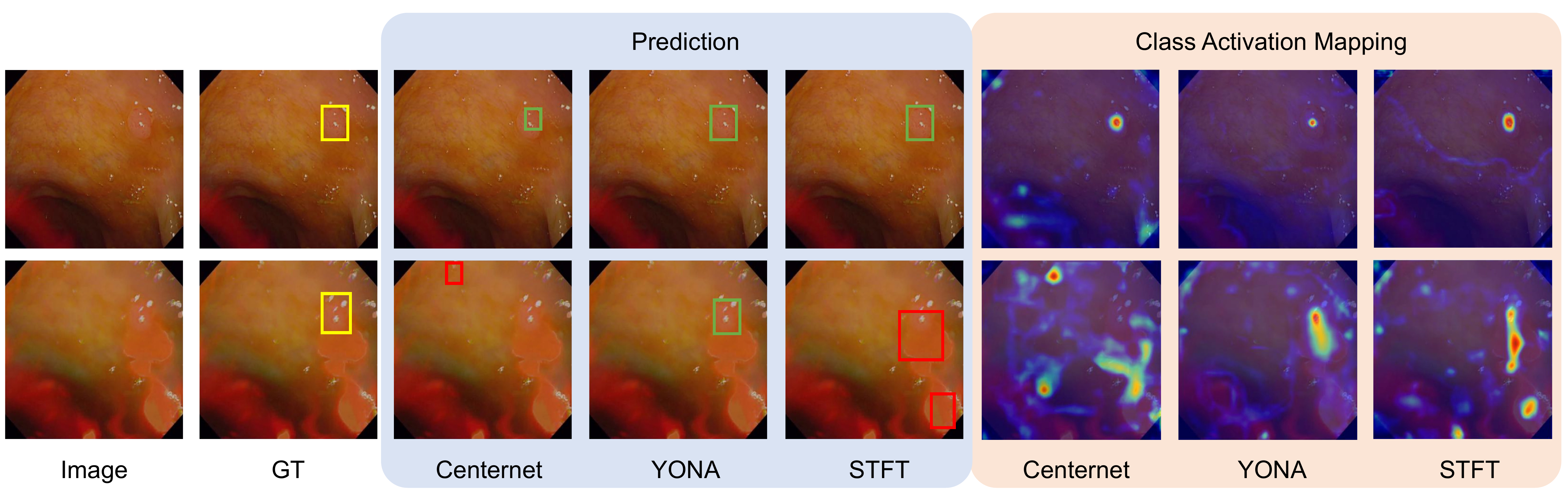}
    \caption{Illustration of attention regions and predictions of different methods on two adjacent frames. Image-level detectors lack the ability to extract useful context information on blurred images, leading to false positive results. Thanks to the proposed foreground and background alignments, our YONA can fuse the context information from the adjacent frame and obtain valid predictions.}
    \label{fig:cam}
\end{figure}

%
%